# Knowledge Graph Construction in Power Distribution Networks

Xiang Li[1], Che Wang[1], Bing Li[1], Hao Chen[2], Sizhe Li[2]
1. Nanjing Power Supply Bureau, Nanjing, Jiangsu, China, 210000
2. Nanjing University Of Finance & Economics, Nanjing, Jiangsu, China, 210000

**Abstract:** In this paper, we propose a method for knowledge graph construction in power distribution networks. This method leverages entity features, which involve their semantic, phonetic, and syntactic characteristics, in both the knowledge graph of distribution network and the dispatching texts. An enhanced model based on Convolutional Neural Network, is utilized for effectively matching dispatch text entities with those in the knowledge graph. The effectiveness of this model is evaluated through experiments in real-world power distribution dispatch scenarios. The results indicate that, compared with the baselines, the proposed model excels in linking a variety of entity types, demonstrating high overall accuracy in power distribution knowledge graph construction task.

## 0    Introduction

The automation and automation of distribution network scheduling is an important part of distribution network automation. In the construction of distribution network dispatching automation and informatization, the use of intelligent virtual dispatchers instead of manual dispatchers for receiving, verifying and sending on-site dispatching information is conducive to reducing the single repetitive workload of dispatchers, improving the automation degree of distribution scheduling and the dispatchers' analytical decision-making efficiency [1-2]. Since distribution network scheduling involves a large number of power equipment and related equipment status and operation information, in recent years, there have been a number of studies using knowledge graph to organize and analyze distribution scheduling information more effectively [3-4].

In the research related to electric power knowledge graph, many studies have explored the entity linking of electric power text. Literature [5] constructed a knowledge graph of grid scheduling automation system for fault analysis, but it is still necessary to manually locate the entities related to faults in the knowledge graph during fault diagnosis; Literature [6-7] used a direct string matching method to link the text keywords to the corresponding entities in the knowledge graph, but the direct matching method is difficult to understand the semantic information of the text and the entities, and it cannot deal with differences in entity representations of the text (e.g., "knife gate") when applied to the linking of scheduling text entities. However, the direct matching method is difficult to understand the semantic information of text and entities in depth, and when it is applied to the scheduling of text entity links, it can't handle the entity representation differences (such as "knife switch" and "knife switch") and entity discontinuity (such as the entity "Xunyang 298 Line" of "Xunyang 298 Line and Xunbei 299 Line"), and it can't adapt to the text information deviation caused by speech recognition deviation. Literature [8] suggested that cosine similarity can be used to find synonymous entities to deal with the problem of entity representation differences in power texts, but did not explain the calculation method and basis of entity vectors; Literature [9-10] adopts the word2vec method to mine the correlation relationship of word meanings from the actual power texts, which solves the problem of entity expression differences of entities in the process of entity linking, but it is still unable to deal with the problem of entity linking in scheduling texts. unable to deal with the speech recognition bias and entity discontinuity problem of entity linking in scheduling text.

In order to perform deep extraction of text features during supervised learning, many studies in recent years have introduced deep learning neural network-based semantic matching models in entity linking tasks, mainly including recurrent neural networks [17-18] and convolutional neural networks [16,19]. Compared with recurrent neural networks, semantic matching models based on convolutional neural networks are more efficient in applications due to their parallel computing characteristics, and the more representative models include Convolutional Latent Semantic Model (CLSM) [20] and ConvNet [21]. Literature [22], based on ConvNet, proposes to add the information that directly characterizes the literal connection of words to the feature dimension of words by string matching, which enriches the correlation information between the feature matrices of the text to be matched; Literature [23] further proposes Lexical Semantic Feature based Skip Convolution Neural Network (LSM) [24] and ConvNet [25] to provide the lexical semantic features of the text to be matched.which represents the intrinsic semantic connections between the words of the text to be matched by Lexical Semantic Feature (LSF).

When applied to the task of linking text entities for power distribution scheduling, the aforementioned deep learning semantic matching model also needs to be improved in terms of input features and model structure by combining the characteristics of distribution network knowledge graph and distribution scheduling text. To this end, this paper constructs the feature matrices of the actual scheduling text and knowledge graph entities from the three dimensions of semantics, pronunciation and lexicality, and improves the model structure of the LSFSCNN in terms of the feature computation method and the attention mechanism, so as to propose the entity linking method of the distribution scheduling text for the knowledge graph of power distribution network.

## 1 Distribution Grid Knowledge Graph

Since the distribution network scheduling information reported by field personnel usually involves the status of various distribution equipment or related operations, the knowledge graph used for distribution network scheduling mainly consists of entities such as the type and name of the power station, the type and name of the equipment, the type of the equipment status and the type of the operation, and the relationship between them, as shown in Fig. 1.

In Figure 1, the boxes represent the entities. The middle column represents power stations and equipment types, each of which contains multiple specific station or equipment names, as shown in the left column, while each equipment type has a corresponding state or operation, as shown in the right column.

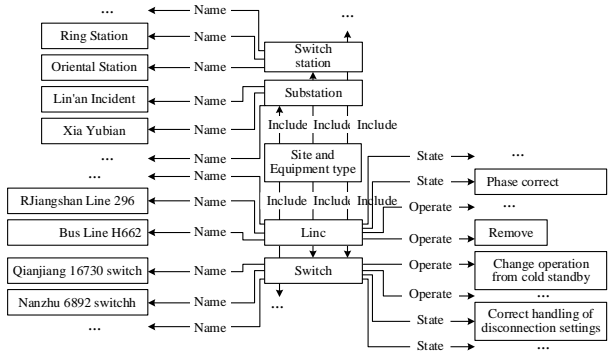

**Fig. 1 Knowledge graph of distribution network**

## 2 Feature extraction of distribution scheduling text

### 2.1 Semantic features of distribution scheduling text

When extracting semantic features of distribution scheduling text, firstly, the scheduling text is divided into words, and then cut into parts with the smallest possible granularity under the premise of maintaining the complete semantic meaning, for example, the name of the switch "Qianjiang 16730 switch" is further cut into "Qianjiang" (representing the name of a place), "16730" (representing the number) and "switch" (representing the device) to obtain more detailed scheduling information and provide more key semantic features for the scheduling text entity links. At the same time, we add the words from the grid ledger and dispatch operation specification into the lexical dictionary, and perform the minimum granularity segmentation and pinyin annotation on the newly added words, so as to improve the accuracy of segmentation and subsequent phonetic notation for the proper nouns such as power stations and equipment, as well as the specialized vocabulary in distribution and dispatching.

After the scheduling text segmentation is completed, word2vec method is used to generate semantic feature vectors for each word [24]. The word2vec includes two structures: Continuous Bag-of-Words (CBOW) and Skip-gram, in which Skip-gram can fully extract the semantic features of low-frequency words, and is more suitable for power distribution scheduling texts containing a large number of low-frequency words such as place names and numbers. Skip-gram structure assumes that the scheduling text. There are $V$ words in this corpus (excluding repeated words), and the semantic features to be generated areThe quantity is $D$ dimension, and the window size is 2 (that is, the first 2 words and the last 2 words of the head word are predicted respectively). Dispatch the one-hot vector $w_i(V)$ of the i-th word (head word) of a sentence; ($V$-dimension) input model, which passes through the input matrix $T_1$ ($V$ row and $D$ column), output matrix $T_2$ ($D$ row and $V$ column) and softmax function, the maximum value of the vector is set to 1, and the rest is set to 0, so as to obtain one-hot vector predictions $w'_{i-2}$、$w'_{i-1}$、$w'_{i+1}$ and $w'_{i+2}$ ( all in $V$ dimension) of four words in the context, and compare their differences with the real one-hot vector, and thenThe parameters of $T_1$ and $T_2$ are trained by back propagation algorithm. After the training is completed each row of the input matrix $T_1$ represents a $D$-dimensional semantic feature vector of a word in the corpus.

### 2.2 Pronunciation features of distribution scheduling text

Due to the fact that on-site staff often speak Mandarin with an accent, coupled with the difficulty of recognizing proper nouns, it is easy to make textual errors on proper nouns when converting on-site input speech into text, such as recognizing the "City Roundabout Station" as the "Flooded City Station". At the same time, the problem of accent may lead to a big difference between the actual pronunciation and the correct pronunciation of some words, not even limited to the more common differences such as flat tongue and warped tongue, front and back nose, and so on.

Therefore, pinyin2vec is constructed by using word2vec for reference to explore the potential connections between words with different pronunciations. Pinyin2vec method is still based on Skip-gram structure, but because accent usually affects the pronunciation of a single word, the pronunciation feature vector of each word in distribution dispatching text is generated first. Suppose that the word with the largest number of words in the dispatch text corpus contains $M$ words, and the pronunciation of each word to be generated is special. If the eigenvector is $D$-dimension, the dimension of the pronunciation eigenvector of each word is

$$C = \lfloor D/M \rfloor \quad (1)$$

Where $\lfloor \ \rfloor$ means rounding down. After the dimension $C$ of the feature vector is obtained, the pronunciation feature vector of a word can be trained by the model, and the one-hot vector of pinyin of each word is input, and the output is the pinyin prediction of four words before and after each word. After the training, the vector of pinyin of each word is obtained, and words with the same pronunciation have the same vector. Finally, the first $C \times N$ dimension of the pronunciation feature vector of a word containing n words is formed by splicing the pronunciation feature vectors of these $N$ words, in which the $(n-1) \times C + 1$ to the $n \times C$ dimension. The feature value represents the pronunciation feature vector ($n = 1, 2, \cdots, N$) of the $c$ word. The remaining $D^- C \times N$ dimension eigenvalues are taken as 0, so that the pronunciation eigenvectors of each word in the distribution dispatching text are obtained based on pinyin2vec method.

**2.3 Lexical features of distribution scheduling text**

There is a relatively fixed pattern in the expression mode of the distribution scheduling text, such as the power equipment is often expressed in the mode of place name + string + noun such as "Beishe 47010 switch" and "Tangxing G224 line", and the equipment state switching is often expressed in the mode of "change from hot standby to cold standby". Equipment state switching is often expressed in the pattern of preposition + verb such as "from hot standby to cold standby" and "from trip to deactivation". Therefore, some semantic information of a word can be inferred through its contextual lexical combination, for example, through the combination of "noun+preposition+noun+verb+noun", it can be inferred that the noun in the middle may be the word indicating the state such as "hot standby", which indicates that contextual lexical properties can be used to represent some features of the word, collectively called "hot standby". This means that contextual lexical properties can represent part of the word's characteristics, collectively called lexical features.

The generation of part-of-speech features is based on the model below. Hypothetical scheduling text. This corpus consists of $V$ words (excluding repeated words) and $V'$ parts of speech (excluding repetition)Complex part of speech), the dimension of the part of speech feature vector to be generated is $D$ dimension, and dispatch the one-hot vector $w_i$ of the $i$-th word (head word) of a sentence; ($V$-dimension) After the model is input, it is multiplied by the input matrix $T'_1$ ($V$ rows and $D$ columns) and the output matrix $T_2'$, in turn. Different from $T_2$, the part-of-speech feature represents the central word based on part-of-speech rather than semantics, the model output corresponds to part-of-speech, and the dimension of the output matrix $T_2'$ should correspond to the number of parts of speech, so t is $D$ rows and $V$ columns. Multiplied by $T_2'$, and then transformed by softmax function and vector maximum value set to 1, the part-of-speech one-hot vector prediction is obtained. Compared with the previous model, the prediction result of the current model includes not only the part-of-speech one-hot vectors $p'_{i-2}$、$p'_{i-1}$、$p'_{i+1}$ and $p'_{i+2}$ of the context (all in $V'$ dimension), but also the part-of-speech one-hot vector p'($V'$ dimension) of the head word, because the part-of-speech of the head word is also part of the part-of-speech combination. Finally, compare the difference between the prediction result of the part-of-speech one-hot vector and the real part-of-speech one-hot vector, and then train the parameters of $T'_1$ and $T'_2$ through the back propagation algorithm. After the training, each row of the input matrix $T'_1$ represents the $D$-dimensional part-of-speech feature vector of a word in the corpus.

## 3 Distribution scheduling text entity linking method

### 3.1 LSFSCNN-based distribution scheduling information matching model

In this paper, based on LSFSCNN, we propose a distribution scheduling information matching model, where the red part is the improvement for the input features of the scheduling text and the blue part is the improvement for the convolution process.

First, the feature matrices of the knowledge graph entities and the dispatching text are generated separately, with each row representing the feature vector of a word. Unlike LSFSCNN, which only transforms the text into a single-layer matrix according to semantic features, this paper transforms the text into a 3-layer matrix representing semantic features, pronunciation features and lexical features described in Section 2, in order to mine the multi-dimensional information of the text by combining the distribution scheduling features. Meanwhile, in LSFSCNN, each word feature vector contains one LSF dimension, which is used to represent the semantic link between two texts to be matched, and its calculation method is as follows

$$\text{vecLSF}(\text{word}_{1j}) = \lceil (1 - \max(0, \max_{1 \leq k \leq ln_2} \cos \langle \text{vec}_{1j\ 2k} \rangle)) \times t \rceil \quad (2)$$

Where: $LSF$ function represents the value of $LSF$ dimension of a word; $\text{word}_{1j}$ is the $j$-th word of text 1; $\lceil \ \rceil$ is an upward rounding symbol; $\boldsymbol{vec}_{2k}$ is the feature vector of the $k$-th word of text 2; $<>$ means to take the included angle of two vectors; $len_2$ is the number of words in text 2; $t$ is the upper limit of LSF value, which is taken as 10. In this model, LSF of each feature matrix not only represents the relationship between feature vectors, but also introduces a new dimension to directly represent the relationship between words in literal, pronunciation and part of speech. The calculation method of new dimension representing literal connection in LSF of semantic feature matrix is as follows

$$\text{Lit}(\text{word}_{1j}) = \min_{1 \le k \le len_2} ED(\text{char}_{1j}, \text{char}_{2k}) \quad (3)$$

Where: the $Lit$ function represents the literal contact eigenvalue of a word; The $ED$ function means to find the editing distance of two sequences; $char_{1j}$ is the word sequence of the $j$ th word in text 1, for example, the word sequence of "overhaul" is {overhaul}; $char_{2k}$ is the word sequence of the $k$-th word of text 2. The calculation method of the new dimension representing pronunciation connection in $LSF$ of pronunciation feature matrix is as follows

$$\text{Pron}(\text{word}_{1j}) = \min_{1 \le k \le len_2} ED(\text{pinyin}_{1j}, \text{pinyin}_{2k}) \quad (4)$$

In the formula: $Pron$ function represents the characteristic value of pronunciation connection of a word; $pinyin_{1j}$ is the phonetic sequence of the $j$-th word in text 1, for example, the phonetic sequence of "overhaul" is {jian3xiu}; $Pinyin_{2k}$ is the phonetic sequence of the $k$-th word of text 2. The calculation method of the new dimension representing the part-of-speech connection in LSF of the part-of-speech feature matrix is as follows

$$\text{Part}(\text{word}_{1j}) = \min_{1 \le k \le len_2} \text{Sim}(\text{pos}_{1j}, \text{pos}_{2k}) \quad (5)$$

Where: the $Part$ function represents the characteristic value of the part-of-speech connection of a word; The $Sim$ function is equal to 0 when the two parts of speech in brackets are the same, and equal to 0.3 when the two parts of speech are nominal verbs and nouns, nominal verbs and verbs, nominal words and nouns, nominal words and adjectives, indicating that the two parts of speech are related to some extent, and the rest are equal to 1; $pos_{1j}$ is the part of speech of the j th word in text 1; $pos_{2k}$ is the part of speech of the $k$-th word in text 2.

Then, the wide convolution method is used to convolution the feature matrix of the knowledge map entity and the scheduled text. Each convolution window is set to three layers according to the number of layers of the feature matrix. As shown in Figure 2, when the conv-olution window completes a convolution at a certain position of the entity feature matrix of the knowledge map, three convolution results representing semantic, pronunciation and part-of-speech features will be generated to form a three-dimensional vector $r_1$. Because the importance of these three types of features in scheduling information matching is different, this paper introduces attention vector $a_1$, and replaces the sum of the three convolution results of r1 with the inner products of vectors $r_1$ and $a_1$ to obtain the comprehensive convolution result. The three-dimension values of $a_1$ can be learned automatically during model training. Similarly, attention vector $a_2$ is introduced into the feature matrix of scheduling text, and the inner product of vectors $r_2$ and $a_2$ generated by convolution is used as the comprehensive convolution result. After the convolution window is convolved at multiple positions, the comprehensive convolution results of all positions will form a comprehensive convolution vector, but in the actual convolution process, multiple convolution windows are usually used to perform convolution operations at the same time to form multiple comprehensive convolution vectors.

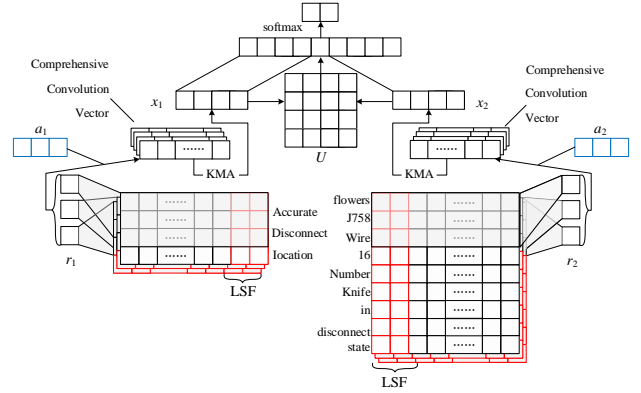

**Fig. 2 Matching model of distribution scheduling information**

Finally, like LSF-SCNN, k-max average pooling (KMA) is performed on the comprehensive convolution vector to generate the knowledge map entity representation vector $x_1$ and the scheduling text representation vector $x_2$, and then calculate the similarity score $x_{\text{sim}}$ with the similarity matrix $U$, as shown in formula (6).

$$x_{\text{sim}} = x_1 U x_2^{\text{T}} \quad (6)$$

After splicing $x_1$, $x_{\text{sim}}$ and $x_2$ into a joint vector, the $softmax$ classifier is used to predict whether the knowledge map entity matches the scheduled text.

### 3.2 Entity Link Method for Distribution Network Knowledge Map

1) segmenting, phonetic notation and part-of-speech tagging all entities of the knowledge map in advance, and

forming a three-layer feature matrix except LSF according to the semantic, pronunciation and part-of-speech feature vectors of each word, and then storing according to the relationship type (name, state and operation) of the pointing entities;

2) After inputting the distribution dispatching text to be matched, the text is segmented, phonetic and part-of-speech labeled, and then the LSF dimension of each entity and the three-layer feature matrix of the dispatching text are generated according to the method in section 3.1;

3) Using the distribution scheduling information matching model in Figure 2, the scheduling text is matched with all entities pointed by different relationship types in turn, so that the scheduling text is linked to all successfully matched entities. After the entity link of the dispatching text is completed, the dispatching text can be analyzed with the help of the knowledge map. For example, by comparing whether the dispatching operation text formed by the voice recognition of the field personnel is consistent with the entity linked to the operation ticket text in the knowledge map, the correctness of the on-site repetition can be judged.

## 4 Experimental Analysis

### 4.1 Experimental data and evaluation index

We use 40,000 distribution dispatching texts collected from Hangzhou Power Supply Company, which were generated by recognizing the voice of field personnel, were all repetitions or reports about dispatching operation or equipment status. Examples of the texts are shown in Table 1. A 50% cross-validation method was used to carry out the experiment, and 40,000 scheduled texts were randomly divided into 5 copies, of which 4 copies were used as training sets and the remaining one as test sets in turn. In each round of experiments, all the training texts are used as the training corpus of semantic, pronunciation and part-of-speech feature vectors, and the training texts and the corresponding knowledge map entities are used as positive sample pairs for the training of distribution dispatching information matching model. Since each dispatching text corresponds to about 3.5 entities on average, the training set of each round of experiments contains about 112,000 positive sample pairs. At the same time, for each training text, seven entities that do not correspond to it are randomly selected in the knowledge map to generate negative sample pairs, and the ratio of positive and negative sample pairs is about 1:2.

After the training, all the scheduling texts in the test set are linked with entities, and the overall accuracy of entity linking of scheduling texts, acc, is calculated, as well as the accuracy of entity linking of three types of

Table 1 Examples of distribution dispatching text

| Distribution dispatching texts | Corresponding entities |
|---|---|
| Linn Creek Line 3170 is indeed in cold standby. | Linxi Line 3170; cold standby indeed |
| Guangfu transformer:.Pull open Sanyuan D45P switch1 | Guangfu transformer: pull open Sanyuan D45P switch. |
| Phase verification on Wulong 50526 line 0# switch. | Wulong 50526 line 0# switch; phase check complete. |
| Phase check complete, phase correct | Phase check complete; phase correct |
| Marine Station2:10kVLine 87276 is under hot standby | Hanghai Station; Huahang 87276 line; by hot standby |
| Non-automatic rerouting for maintenance | Non-automatic rerouting for maintenance |
| Refinery substation, Transportation Bureau switch #1217, both sides | Refinery substation; Transportation Bureau switch #1217. |
| Phase verification complete. Correct phase and phase sequence. | Phase verification complete; phase position correct; phase sequence correct. |

names, states and operations, $acc_{name}$, $acc_{state}$ and $acc_{operate}$. The $acc$ is calculated as

$$acc = \text{count}(text_{correct})/\text{count}(text) \quad (7)$$

Where: the $count$ function indicates the amount of a certain text; $text_{correct}$ indicates the text that the entity link is completely correct; Text represents all the text of the test set. $acc_{name}$ is calculated as follows

$$acc_{name} = \text{count}(text_{name\_correct})/\text{count}(text_{name}) \quad (8)$$

In addition, the entity link time of all texts in the test set is counted, and the average entity link time $t$ of each text is calculated. The graphics card model used in the experiment is NVIDIA TITAN V.

### 4.2 Experimental Results and Analysis

In the experiment, the context window size of the model for generating semantic, pronunciation and part-of-speech feature vectors of dispatching text in this model is 2, and the vector dimension is 50. The number of convolution windows for convolution operation of entity or text feature matrix in distribution dispatching information matching model is 100, and the height is 5 (i.e. five words are convolved at a time), and the $k$ value of KMA is 2.

At the same time, in order to compare the differences in name, status and link accuracy of operating entities among the models, the statistical results of $acc_{name}$、$acc_{state}$ and $acc_{operate}$ are plotted as a column chart. From Table 3, it can be seen that the overall accuracy of this model in the linking task of distribution dispatching text entities and the linking accuracy of all kinds of entities are higher than other methods, and the overall accuracy is over 90%, which shows the effectiveness of this model in extracting the characteristics of dispatching text and improving the dispatching information matching model. Although the model in this paper takes longer than the comparison models, the average link time of each scheduled text is less than 0.5 s, which has good practicability. By further analyzing the experimental results in Table 3, the following conclusions can be drawn:

Table 2 Control models of entity linking

| Distribution dispatching text | correspondent entity |
| --- | --- |
| Control model 1 | Direct matching, if the text contains entity word sequence, the matching is successful. |
| Control model 2 | Match by word. If the text contains every word of the entity, the match is successful. |
| Control model 3 | LSF-SCNN model |
| Control model 4 | This model omits text pronunciation features. |
| Control model 5 | This model omits the part-of-speech features of the text. |
| Control model 6 | The new dimension of LSF is omitted in this model. |
| Control model 7 | This model omits the attention vector. |

Table 3 Overall accuracy of entity linking of all the models

| Model | acc/% | t/s |
| --- | --- | --- |
| Control model 1 | 59.17 | 0.007 |
| Control Model 2 | 62.97 | 0.012 |
| Control Model 3 | 72.09 | 0.390 |
| Control Model 4 | 80.09 | 0.438 |
| Control Model 5 | 82.62 | 0.435 |
| Control Model 6 | 86.77 | 0.456 |
| Control Model 7 | 86.72 | 0.468 |
| Model of this paper | 91.94 | 0.471 |

Compared with the semantic matching models based on convolutional neural network (control model 3 to model 7 and this model), the overall accuracy and each type of entity linking accuracy of the control model 1 with direct matching and the control model 2 with word-by-word matching are significantly lower, because it is difficult to understand the semantic information of the dispatch text by literal matching, and when there are textual discrepancies caused by the expression differences, speech recognition bias and entity discontinuities between the dispatch text and the entities of the knowledge graph, it is difficult to understand the semantic information of the dispatch text by literal matching, and when there are textual discrepancies caused by the expression differences, speech recognition bias and entity discontinuities. When the scheduling text and the knowledge graph entities have textual differences caused by expression differences, speech recognition bias and entity discontinuity, it is difficult to link the scheduling text to the corresponding entities. Compared with direct matching, word-by-word matching can deal with the problem of entity discontinuity in some scheduling texts, such as recognizing the state entity "phase correct" in "phase and phase sequence correct", but it still cannot solve the differences in entity expressions (e.g., "reclosing" and "resuming"). However, it still cannot solve the problems of differences in entity expressions (e.g., "reclosing gate resumption" and "reclosing gate restoration") and voice recognition bias, so the improvement of entity linking accuracy is limited.

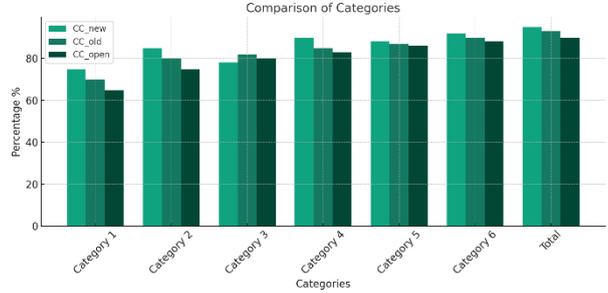

Fig. 3 Each type of entity linking accuracy

Comparing the experimental results between this model and LSFS-CNN (control model 3), we find that this model has a greater advantage in $acc$ and all kinds of entity link accuracy. The main reason is that the model in this paper, based on LSFS-CNN, extracts more relevant features for the distribution network knowledge graph entities and distribution scheduling text, including the introduction of articulation features to solve the voice recognition bias caused by the accent problem of field personnel, and the introduction of lexical features to discover the expression pattern of distribution scheduling language, thus improving the model's adaptability to the text entities of distribution scheduling for the knowledge graph of the power distribution network;

Comparing the acc of this paper's model with that of the control model4, it is found that the failure to include dispatch text pronunciation features has a greater impact on the $acc$. Further analyzing the accuracy of entity linking for various types of entities, it is found that the low $acc$ of the control model4 is mainly due to the low $acc_{name}$. Due to the problem of field noise and personnel accent, and the difficulty of recognizing proper nouns, when the pronunciation of proper nouns is relatively close to that of common words, it is easier to make textual errors in the scheduling text after speech recognition, such as recognizing "Huaming Station" as "Invention Station", "Shangshu Transformation" as "Invention Station", and "Huaming Station" as "Invention Station". For example, "Hua Ming Station" is recognized as "Invention Station", and "Shangshu Station" is recognized as " Shangshu Station", etc. Therefore, the effect of not including the pronunciation features of scheduling text on $acc_{name}$ is larger. Compared with proper nouns, the text indicating status and operation contains more common words, and the difficulty of speech recognition is smaller, and the reason for errors is mainly the misrecognition of a small number of distribution scheduling vocabulary, such as recognizing "nuclear phase" as "combined item", and "de-listing" as "well-known", and so on. Therefore, the $acc_{state}$ and $acc_{operate}$ of the control model 4 are relatively high.

Comparing this model with the control model5, it can be found that the lexical features of the scheduling text also have a greater impact on acc, but the main reason for the low acc is the low $acc_{state}$ and $acc_{operate}$. When some non-useful expressions of state or operation appear in the scheduling text, it is usually difficult to dig out the meaning of the words only by semantic features, but the meaning can be inferred by the lexical combination of the context of the words. For example, "reclosing gate / already / in / signal" is usually expressed as "reclosing gate/confirmation/location/signal" ("/" denotes the position of participle), the difference between the two state expressions is 2 words, but the lexical combination of both is "noun/adverb/verb/noun", which can be deduced that both are for the state of the equipment expression mode, combined with the same real words such as "reclosing" and "signal" can be used to match the two. At the same time, the name entity also exists more place names, strings and other lexical properties that are different from the general words, so adding scheduling text lexical features also plays a role in improving the accuracy of the name entity link $acc_{name}$.

Comparing this paper's model with the control models6 and 7, it can be found that adding new dimensions and attention vectors to the LSF in the distribution scheduling information matching model can further improve the model's $acc$ and link accuracy of various entities. The new dimension of LSF directly represents the relationship between words in literal, pronunciation and part of speech, which provides an effective supplement to LSF's original method of indirectly reflecting the relationship between words through cosine similarity of feature vectors. Attention vector can dynamically adjust the attention weights of semantic, pronunciation and part-of-speech features through automatic learning in model training, so as to reflect the difference of importance of different features in information matching, thus improving the accuracy of entity link of the model.

## 5 Conclusion

The experimental results show that the model in this paper can achieve a high level of accuracy in the overall accuracy of scheduling text entity linking and the linking accuracy of various types of entities, and the construction of scheduling text features and the improvement of the LSFSCNN structure are conducive to improving the application effect of entity linking. Due to the high reliability requirements of power distribution scheduling, the subsequent work needs to further improve the entity linking accuracy, and there are two main research directions: ; and second, the improvement of the practical application methods, such as Designing a question-and-answer mechanism, when the confidence level of the entity link results output from the model is not high, the virtual dispatcher will ask questions to the field personnel to confirm the relevant information and ask the field personnel to repeat, so as to improve the reliability of the entity link.

**References:**

[1] Jain, T., Ghosh, D., Mohanta, D. K. "Augmentation of Situational Awareness by Fault Passage Indicators in Distribution Network Incorporating Network Reconfiguration." *Protection and Control of Modern Power Systems*, vol. 4, no. 4, 2019, pp. 323-336. DOI: 10.1186/s41601-019-0140-6.

[2] Yang, Liqing. "Voice Man-Machine Interaction and Its Application in Intelligent Scheduling." Ph.D. Dissertation, Shandong University, 2013.

[3] Yang, Liqing. "Application of Human-Machine Voice Interaction in Intelligent Scheduling." Ph.D. Dissertation, Shandong University, 2013.

[4] Gao, Zepu et al. "Identification Method of Low Voltage Distribution Network Topology Based on Knowledge Graph." *Power System Protection and Control*, vol. 48, no. 2, 2020, pp. 34-43.

[5] Yu, Jianming et al. "Construction and Application of Knowledge Graph in the Field of Intelligent Control." *Power System Protection and Control*, vol. 48, no. 3, 2020, pp. 29-35.

[6] Li, Xinpeng et al. "Construction and Application of Knowledge Graph in Dispatch Automation System." *China Electric Power*, vol. 52, no. 2, 2019, pp. 70-77.

[7] Li, Xinpeng et al. "Construction and Application of Knowledge Graph in Electric Power Dispatch Automation System." *Electric Power*, vol. 52, no. 2, 2019, pp. 70-77.

[8] Ji, Yuan et al. "Construction Method of Semantic Search System in Power Field." *Computer Systems Applications*, vol. 25, no. 4, 2016, pp. 91-96.

[9] Xiong, Siheng, et al. "TILP: Differentiable Learning of Temporal Logical Rules on Knowledge Graphs." The Eleventh International Conference on Learning Representations. 2022.

[10] Wang, Y. et al. "Construction and Application of Knowledge Graph in Power System Full-Service Data Center." *IOP Conference Series: Materials Science and Engineering*, vol.

452, 2018, p. 19.
- [11] Liu, Ziquan, Wang, Huifang. "Retrieval Method of Power Equipment Defect Records Based on Knowledge Graph Technology." *Automation of Electric Power Systems*, vol. 42, no. 14, 2018, pp. 158-164.
- [12] "Method for Error Recognition of Power Equipment Defect Records Based on Knowledge Graph." *Frontiers of Information Technology & Electronic Engineering*, vol. 20, no. 11, 2019, pp. 1564-1577.
- [13] Han, X. P., Sun, L. "A Generative Entity-Mention Model for Linking Entities with Knowledge Base." *Annual Meeting of the Association for Computational Linguistics: Human Language Technologies*, 2011, Portland, pp. 945-954.
- [14] Zhang, W. et al. "Effective Entity Linking with Initials Expansion, Instance Selection, and Topic Modeling." *International Joint Conference on Artificial Intelligence*, 2011, Barcelona, Spain, pp. 1909-1914.
- [15] Alhelbawy, A., Gaizauskas, R. "Graph Ranking for Collective Named Entity Disambiguation." *Annual Meeting of the Association for Computational Linguistics*, 2014, Baltimore, USA, pp. 75-80.
- [16] Huang, H. Z. et al. "Leveraging Deep Neural Networks and Knowledge Graphs for Entity Disambiguation." *Computer Science*, 2015, pp. 1275-1284.
- [17] Xia, F. et al. "Learning Rank Theories and Algorithms." *International Conference on Machine Learning*, 2008, Helsinki, Finland, pp. 1192-1199.
- [18] Zhang, Guangpeng. "Design and Implementation of a Specific Domain Entity Linking System." Ph.D. Dissertation, Harbin Institute of Technology, 2017.
- [19] Liu, Bin. "Research on Entity Linking Based on Graph Models and Deep Learning." Ph.D. Dissertation, Central China Normal University, 2018.
- [20] Luo, Angen. "Research on Entity Linking Algorithm Integrating Knowledge Graph." Ph.D. Dissertation, Beijing University of Posts and Telecommunications, 2017.
- [21] Sun, Y. M. et al. "Modeling Mention, Context, and Entities with Neural Networks for Entity Disambiguation." *International Joint Conference on Artificial Intelligence*, 2015, Buenos Aires, Argentina, pp. 1333-1339.
- [22] Shen, Y. L. et al. "Convolutional Pool Structure Latent Semantic Model for Information Retrieval." *ACM International Conference on Information and Knowledge Management*, 2014, Shanghai, China, pp. 101-110.
- [23] Severyn, A., Moschitti, A. "Learning to Rank Short Text Pairs with Convolutional Deep Neural Networks." *International ACM SIGIR Conference on Research and Development in Information Retrieval*, 2015, Santiago, Chile, pp. 373-382.
- [24] Severyn, A., Moschitti, A. "Modeling Relational Information in Question-Answer Pairs." [Online]. Available: https://arxiv.org/pdf/1604.01178.pdf
- [25] Guo, J. H. et al. "An Enhanced Convolutional Neural Network Model for Answer Selection." *International Conference on World Wide Web Companion*, 2017, Perth, Australia, pp. 789-790.
- [26] Mikolov, T. et al. "Efficient Estimation of Word Representation in Vector Space." *International Conference on Learning Representations*, 2013, Scottsdale, USA, pp. 1-12.